\documentclass[11pt,a4paper]{article}
\usepackage[hyperref]{eacl2021}
\usepackage{times}
\usepackage{latexsym}
\usepackage{alltt}  %
\usepackage{mathtools}
\usepackage{amsmath}
\usepackage{amssymb}
\usepackage{array,multirow,graphicx}
\usepackage{soul}
\usepackage{latexsym}
\usepackage{graphicx, subfigure}
\usepackage{url}

\newcommand{\commentout}[1]{}
\usepackage{microtype}

\aclfinalcopy 

\title{Representations for Question Answering from Documents with Tables and Text}

\author{Vicky Zayats$^\dagger$\thanks{*Most of the work was done while the author was at the University of Washington.} \quad
  Kristina Toutanova$^\dagger$ \quad
  Mari Ostendorf$^\S$ \\
  $^\dagger$Google Research \quad $^\S$University of Washington \\
  \{\texttt{vzayats, kristout}\}\texttt{@google.com} \quad 
  \texttt{ostendor@uw.edu}\\}

\date{}

\begin{document}
\maketitle
\begin{abstract}
Tables in Web documents are pervasive and can be directly used to answer many of the queries searched on the Web,
motivating their integration in question answering.
Very often information presented in tables is succinct and hard to interpret with standard language representations. On the other hand, tables often appear within textual context, such as an article describing the table. Using the information from an article as  additional context can potentially enrich table representations. In this work we aim to improve question answering from tables by refining table representations based on information from surrounding text. We also present an effective method to combine text and table-based predictions for question answering from full documents, obtaining significant improvements on the Natural Questions dataset.

\commentout{
combine text and table-based models
First, we introduce a training scheme that improves the overall accuracy of the question answering on tables using a subset of the Natural Questions dataset \cite{47761}. Secondly, we propose a new approach that propagates information from the article surrounding the table to the relevant parts of the table, further updating the table representation.
}
\end{abstract}

\section{Introduction}
Tables are a common type of information representation used across the Internet. With billions of search queries a day,\footnote{https://www.internetlivestats.com} question answering on tables is an important task that translates into a large number of search queries every second about information present in tables.
In general, research on Question Answering (QA) can be categorized in terms of the resources that are used in answering the question: text documents (often referred as \textit{unstructured} text in the literature), tables, or a structured knowledge base (KB). In our work we are interested in the combination of text-based and structured resources for question answering, particularly articles that contain both tables and text. This is a natural next step for question answering on tables, in that most tables are embedded in documents that discuss them, creating the challenge of determining whether the answer is in the text or the table (if anywhere). In addition, very often information presented in tables is compact and 
abbreviated.
The associated text can potentially provide rich context that can be used to enhance the representation of the table for more robust question answering. 

The main focus of this paper is to investigate how to improve question answering on documents that contain both text and tables. While recently there has been a lot of interest in reading comprehension for both text and tables,  little research has been done in combining the two sources of information. The only prior study we are aware of is by \citet{chen2020hybridqa} who introduced a new dataset for multi-hop QA over tabular and textual data. In their work, the authors heavily rely on the assumption that the questions would be unanswerable if either text or table information is missing. 
Here we investigate a more realistic scenario of naturally occurring questions, where the answer can be found in either text, tables, both, or none. We evaluate our approach on the Natural Questions corpus \cite{47761} which consists of real anonymized queries issued to the Google search engine and corresponding Wikipedia articles, simulating a real use case of such a system. 

Prior work on the Natural Questions dataset has treated text and tables uniformly, linearizing tables and representing them and text segments using the same contextual token representations (for example, starting from pre-trained transformers~\cite{vaswani2017attention} like BERT~\cite{devlin2018bert}). However, representations developed for text are sub-optimal for tables, since they do not account for the special relationships between table cells, defined by the row and column structure.   In this work, we extend the BERT architecture to account for inter-cell relationships in tables. This approach is motivated by Graph Neural Networks with a transformer \cite{shaw2018self} and is closely related to the one in \citet{muller2019answering}. In our work, we pretrain parameters for the new relationships using a large table corpus extracted from Wikipedia \cite{bhagavatula2013methods}. 

\commentout{In order for question answering to work effectively on tables, it is important to model a structure presented in the table. Text in tables have special relations, such as relations between cells and corresponding row and column headers, or nearby table cells. kt -- I think this is redundant}

In addition, we present a novel approach that refines table representations by attending to related representations of text in the surrounding article. This allows information to propagate from the text to table elements, improving the ability of the model to interpret tables and find answers in them. 

\commentout{allows learning a better table representation from the text of the article to provide a more robust table representation. This mechanism allows to propagate an information from the context (textual part of an article) to the table entries by updating the corresponding token embeddings. }

Overall, the main contributions of our work are three-fold. First, to the best of our knowledge, this is the first work that investigates how to effectively combine text-based and table-based approaches in a setting where it is unknown which, if any, of these modalities contains an answer to a question. Second, we introduce a novel mechanism to enrich table representations based on text surrounding the table, which improves the performance of a model for question answering from tables. Finally, this is the first model that uses the Natural Questions corpus for question answering on tables, improving the baseline in \citet{alberti2019bert} that does not distinguish between tables and text.

\section{Related Work}

Most work on QA with tables prior to BERT involves first converting the table to a Knowledge Graph (KG) where cell entries are entities with row/column relations, then using entity linking to identify spans in the question that match an entity in the knowledge graph, and finally parsing the question to generate a SQL query using some variant of a sequence-to-sequence model \cite{krishnamurthy2017neural}. Due to the advances in contextualized word embeddings, more recent work proposed a modification of the BERT transformer architecture to be used for representing tables.  \citet{hwang2019comprehensive} proposed the usage of additional \url{[SEP]} tokens between headers of the table to make a BERT model more suitable for the tables. Recently, \citet{yin2020tabert} introduced a pretraining procedure for joint representation of tabular data paired with an utterance, where the approach is to linearize the structure of tables to be compatible with a BERT model.  Our approach for table encoding is most similar to that of \citet{muller2019answering}, where the authors generalized the BERT architecture similarly to \citet{shaw2018self} with new types of relations to encode table-specific relationships. The main differences between our table representation and \citet{muller2019answering} is that in our representation we use 5 types of relations, cell-column, cell-row, in-cell, cross-column and cross-row (more details in Section \ref{sec: qa_methods_table_rels}), while in their work the authors use cell-column and cell-row relations only for the table representation, but in addition use question-cell relations for marking matches between tokens in the question and corresponding cell values. 
Finally, the two most recent works on table representation learning, TaPaS \cite{herzig2020tapas} and GraPPa \cite{yu2020grappa}, also use pretraining on the Wikitables dataset \cite{bhagavatula2013methods} that we use in our work. Therefore, our table representations based on transformers and our pretraining method are comparable to those in recent and concurrent work.  

Leveraging tables is a hard problem. However, most studies on table-based QA omit an important additional information source: the text in the article discussing the table. Prior attempts at integrating a KB and text use early fusion of document text and KG information \cite{sun2018open}, where they integrate text and a KG sub-graph in a single graph, from which an entity is selected to answer the question. Structured KGs are often easier to interpret than tables, which have a wide variety of possible schemas. 
InfoTabS \cite{gupta2020infotabs} introduced a dataset for the natural language inference task based on premises that are tables, where the authors 
explore multiple table representations, including a key-value approach and linearized representations with table rows corresponding to ''sentences.''
Hypothesis representations are calculated separately.
Recently, TaBERT \cite{yin2020tabert} introduced a joint table-utterance representation approach, where a table row is concatenated with a short text utterance, such as the query in question answering, and passed as an input to a BERT-based model. 
Such an approach relies on the initial table representation to select the table rows most relevant to the query.
In contrast, we enrich the table representation 
using an attention mechanism with the representations of the most relevant parts of the context of the article in which the table appears. 


The Natural Questions is a large corpus that contains real user queries along with their corresponding Wikipedia articles, which may or may not contain an answer anywhere in the article. \citet{alberti2019bert} provided a BERT-based baseline that treats both table and text segments like text: a sequence of tokens with word and position embeddings. Recently, \citet{liu2020rikinet} improved this baseline by using dynamic dual-attention over paragraphs and cascade answer predictor. In another direction, \citet{ravula2020etc} used an extended transformer architecture that models extra-long documents with limited propagation of information among different segments. All three approaches did not distinguish between text and table input, treating tables as text while not taking into account table structure. To the best of our knowledge, this is the first work that focuses on table-based QA for the real user queries in the Natural Questions corpus, and shows that table-based and overall QA performance can be improved by building on state-of-the-art pre-trained representations of table structure, additionally enriched via attention to related article text. 

\section{Methods}
\label{sec: qa_methods}

This section describes two methods that improve table representations for QA from tables: an extension of the BERT architecture for table encoding to better capture the relationships between table elements,
and a mechanism that incorporates related unstructured text of an article as context to further improve the table representation. 
 We then describe an approach for question answering from both text and tables that  combines predictions from text QA and table QA models using a late fusion approach. 
 
\subsection{Table Encoding}
\label{sec: qa_methods_table_rels}
Our table encoding is a generalization of the transformer architecture \cite{vaswani2017attention}, with the self-attention sub-layer extended to incorporate relations between structural components in tables, similar to the one introduced by \citet{muller2019answering}. This approach is motivated by Graph Neural Networks (GNNs) with a transformer \cite{shaw2018self,shaw2019generating} where the authors generalize the transformer by introducing multiple types of relations between inputs. 

\subsubsection{GNNs with a Transformer}
In the original transformer, 
the multi-head self-attention for head $h$ is calculated using query (Q), key (K) and value (V) projections as follows:
\begin{align*}
\label{eq: transformer}
    Att(Q, K, V ) &=\\ softmax&(\frac{(W_h^QQ)^T(W_h^KK)}{\sqrt{d_k}})W_h^VV
\end{align*}
where $W_h^Q$, $W_h^K$ and $W_h^V$ are learned parameters, 
and $d_k$ is the query/key dimension. By calculating the dot product between query and key projections, a transformer captures the interaction between each pair of inputs $x_i$ and $x_j$ (e.g. wordpieces in BERT) at positions $i$ and $j$ for $0\leq i,j\leq N$. This interaction can be generalized to account for relation type $t$ between $x_i$ and $x_j$ by biasing the key projection using $r^t_{ij}$:
$$    s^h_{ij} = \frac{(W_h^Qx_i)^T(W_h^Kx_j + r^{t,h}_{ij})}{\sqrt{d_k}}$$
and then scaling using softmax across all inputs $0\leq j \leq N$. 
Thus, the standard transformer can be considered as a special case with $r_{ij} = 0$. 
Similarly, the value projection can also be updated with the corresponding relation type represented using the bias term $\rho^t_{ij}$, with the overall attention head $h$ calculated as follows: 
\begin{align*}
    \alpha^h_{ij} &= softmax(\frac{s^h_{ij}}{\sum_j s^h_{ij}}) \\
    w^h_i &= \sum_j \alpha^h_{ij} (W_h^Vx_j + \rho^{t,h}_{ij}) 
\end{align*}
The parameters $r^{t,h}$ and $\rho^{t,h}$ are head- and layer-specific.

\begin{figure*}[t]
\includegraphics[scale=0.4]{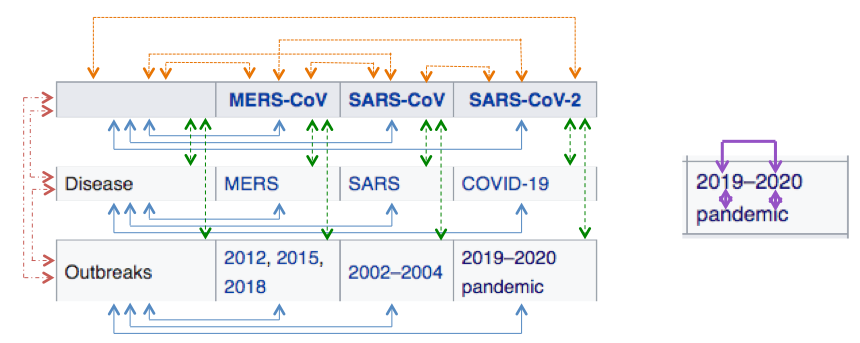}
\centering
\caption{Example of a table encoding with relations: (\textit{green, dashed}) token cell - token column header relations; (\textit{blue, solid}) token cell - token row header relations; (\textit{violet, bold}) in-cell token relations; (\textit{red, dash-dotted}) cross-column header relations; (\textit{orange, dotted}) cross-row header relations.}
\label{fig: qa_table_enc_example}
\end{figure*}

\subsubsection{Table Encoding using GNN}
\label{sec: relations}
Table structure can be encoded using the model described above to account for special relations in the table. Here, we use the following relation types:
\begin{itemize}
  \setlength\itemsep{0em}  \item token cell - token column header relation
    \item token cell - token row header relation
    \item in-cell token relations
    \item cross-column header relations
    \item cross-row header relations
\end{itemize}

For each of these table-specific types of relations we learn different type-specific biases $r^t$ and $\rho^t$ for each layer and head, while for the rest of the relations we use the original BERT configuration with zero bias. Figure \ref{fig: qa_table_enc_example} shows an example of a table with the table relations used in this work.

Our table encoding is similar to the one independently proposed by \citet{muller2019answering} with the main difference of having relations on a token level rather than cell level. Also, in their approach, the authors use only cell-column and cell-row relation, while in our work we also use cross-column header, cross-row header and in-cell relations. The above mentioned paper includes additional relations based on n-gram matches with the question, and special processing of numerical values.


\subsubsection{Pretraining}
\label{sec: pretraining}
The main motivation for applying the original BERT model to table encoding is to use contextualized embeddings that are pretrained on a large amount of data. The new relation biases incorporated as part of the proposed table encoding are randomly initialized, starting from a standard transformer model pre-trained on text. Since these parameters are added for each  layer, they can significantly change the activations of the pretrained model. In order to derive a better initialization point for the additional bias terms, we pretrain a GNN model with the masked LM objective used in BERT on the 
Wikitables dataset \cite{bhagavatula2013methods}, which contains 1.6M tables from English Wikipedia. In order to limit the amount of overfitting on that table set, we freeze all original BERT parameters while updating only the bias terms introduced in the GNN. We tune the model using perplexity on a subset of the Wikitables dataset.  

\subsection{Context-aware Table Representation}
\label{sec: qa_methods_text_aware}

We hypothesize that text in the article of a Web page containing a table can help build an improved representation of the table for QA. Recent work has explored building encoders over large input sequences.
\citet{ravula2020etc} scaled input sequence length to more than 8,000 tokens for the NQ dataset. However, to make the model efficient, encodings of individual text or table segments communicate through single-vector global memories.
Here, we take the approach of using asymmetric attention from table token representations to a small number of relevant text token representations, that are pre-computed independently. 
Our approach is more similar to the handling of prior segment context in Transformer-XL~\cite{dai-etal-2019-transformer}, but relevant context is selected based on word overlap and not contiguity.


The two components of our approach, described next, include the definition of relevant text context for table elements and the mechanism for using contextualized embeddings of the relevant text to enrich the table token representations.


\subsubsection{Table-Textual Context Linking}

Let a table cell that contains a sequence of input tokens 
be defined as $(u_0^t \dots u_{K}^t)$, with the corresponding $s$ sub-word units (wordpiecies in BERT or byte-pair encodings in RoBERTa) 
for the $k$-th word to be defined as $(x_0^{t,k} \dots x_{S_k}^{t,k})$, and let the textual context of the article surrounding the table be defined as $(u_0^c \dots u_{N}^c)$, with the corresponding sub-word units for the $n$-th word be as $(x_0^{c,n} \dots x_{S_n}^{c,n})$. For each word in the table $u_i^t$, we find the corresponding context in the text using the exact match of the lower-case sequence of tokens, starting with the  trigram matches, following with bigram and unigram matches.\footnote{Function words are frequent in multi-word expressions. To avoid exact match of expressions solely consisting of function words,
matching expressions must contain at least one non-frequent word, defined based on the 200 most frequent words in the training set and
the NLTK stop-word list.} For example, a trigram match for a word $u_i^t$ is $\alpha_3(u_i^t) = u_j^c$  if a lower-case expression $(u_i^t,u_{i+1}^t,u_{i+2}^t)$ equals to the lower-case expression $(u_j^c,u_{j+1}^c,u_{j+2}^c)$. For each of the table tokens $u_k^t$ we collect up to 6 corresponding matches from the text, ${u_{s1}^c \dots u_{s6}^c}$, and extract their sub-word embeddings represented by the last layer of pretrained RoBERTa, 
$e(x_0^{c,s1}) \dots e(x_s^{c,s6})$. Then, we stack all the sub-word unit embeddings associated with $u_k^t$ to get a text-aware representation for word $k$ in the table $e(u_k^t) = [e_0(x_0^{c,s1}); \dots; e_r(x_s^{c,s6})]\in \mathbb{R}^{r \times d}$, with $d$ being the size of the hidden RoBERTa embedding and $r$ being the total number of sub-word units corresponding to ${u_{s1}^c \dots u_{s6}^c}$. For simplicity of implementation, we use $r=12$, where we prune any extra word-piecies or use padding for the cases where $r \ne 12$. 

\subsubsection{Enriching Table Representations}

In order to incorporate text context $e(u_i^t)$ into a table representation of $x_s^{t,i}$, we use a slightly modified version of self-attention in the transformer, with the goal of generating an additional text context head, concatenated with the rest of the attention heads 
across multiple transformer layers that would be aware of the broader text context.

In the original transformer,  self-attention is calculated between an input at  position $i$ using a query projection $W^Qx_i$ and the rest of the inputs in  positions $j \in 0 \dots N$ using  key and value projections,  $W^Kx_j$ and $W^Vx_j$, correspondingly. Now,  when calculating a text-aware head for each of the sub-word units $x_i^{t,k}$, we use attention between a query of $x_i^{t,k}$ and all of the sub-word units of the corresponding text context $e(u_k^t)$ using the key and value projections $W_e^Ke(u_k^t)$ and $W_e^Ve(u_k^t)$, correspondingly:
\begin{align*}
    Att(Q_e, K_e, V_e ) &=\\ softmax&(\frac{(W_e^QQ)^T(W_e^Ke(K))}{\sqrt{d_k}})W_e^Ve(V)
\end{align*}

Then, we concatenate the new text-aware head with the rest of the heads in the layer $h$, resulting in a total of $m+1$ heads ($m=16$ for $\text{RoBERTa}_\text{large}$), each of a size $k=64$. 
In addition, we extend the current projection layer from a size $km \times km$ to $k(m+1) \times km$ 
in order to fit the additional head, randomly initializing the additional $k \times km$ parameters. 
For computational efficiency, we incorporate the text-aware representation only at layers 12, 16, 20 and 24.

\subsection{Combining Text and Table Answer Predictions}
\label{sec: methods_text_table}

In question answering, both text and table contexts can be used to support meeting the user information need. Question answering systems should therefore be able to consider both sources of information to present the most suitable answer.
So far we have presented enriched table representations 
that can be used for question answering from tables. We now consider approaches for the full document-level QA task, where an answer may be found in either or none of the two modalities.

Since a text-based QA model would not benefit from the architecture and pre-training extensions for our table representations, we use a standard text-based representation for QA from text. We combine predictions from two separate models for the full document-level QA task.
Specifically, we train a generic model for full article question answering following \cite{alberti2019bert}. This model assigns scores to candidate answers in both text and tables using a standard pre-trained text representation (RoBERTa).
We also train a separate model which uses enriched table representations and pre-training, and focuses on predicting answers in tables. 
The two model predictions are combined
using a late fusion approach detailed below.


\subsubsection{Calculating Prediction and Confidence Scores}
\label{sec: qa_conf}
We follow \citet{alberti2019bert} to define a loss function for training and an answer span prediction method. More specifically, at inference time the scores that correspond to the start and end of a possible answer span are defined as follows:
\begin{align*}
  g(c,s,e) &= f_{start}(s,c; \theta) + f_{end}(e,c; \theta) \\&- f_{start}(s=[\text{CLS}],c; \theta) \\&- f_{end}(e=[\text{CLS}],c; \theta)      
\end{align*}
where $c$ is a context of 512 sub-word unit ID's (including question and document tokens),
$s, e \in \{0, 1, . . . , 511\}$ are inclusive indices pointing to the start and end of the target answer span, $\theta$ is our model parameters, and  $f_{start},f_{end}$ are two different outputs derived from the last layer of our model using linear projections. Following this work, the \url{[CLS]} token is used at training time to
predict no answer instances, making $g(c,s,e)$ the log-odds of the likelihood of an answer span and the \url{[CLS]} span. 
All the contexts from each document are scored and document spans $(s, e)$ are ranked to return the highest scoring span 
that does not exceed 30 tokens. We denote the highest scoring span for the generic model as $g_c$, and the highest scoring span for table model as $g_t$. 

In addition to the prediction score, we also calculate a confidence score $\kappa$ for each span:
\begin{align*}
  \kappa(c,s,e) &= \log(\frac{\exp(f_{start}(s,c; \theta))}{\sum_{s'}\exp(f_{start}(s',c; \theta))}) \\&+  \log(\frac{\exp(f_{end}(e,c; \theta))}{\sum_{e'}\exp(f_{end}(e',c; \theta))})
\end{align*}

\subsubsection{Combining Predictions}
\label{sec: comb_preds}
To combine the scores of the generic model
$g_{c}$ and the table model $g_{t}$, we use grid search on three parameters: a scaling factor $\alpha$, a bias $\beta$, and a confidence threshold $\gamma$ associated with the confidence of the table prediction $\kappa_{t}$:
\begin{align*}
 g =
  \begin{cases}
    \max\limits_{t,c}  (g_{c}, \alpha \cdot g_{t} + \beta)  \quad &\text{if}\,  g_{c}^{l} \in \text{t},\kappa_{t} > \gamma\\
    g_{c} \quad &\text{otherwise}
  \end{cases}
\end{align*}
where $g_{c}^{l}\in \text{t}$ indicates whether the long answer span that is predicted by the generic model points to a table. 
The search for parameters $\alpha$, $\beta$, and $\gamma$ is done on a validation set.
\section{Experiments}

We evaluate our model on the Natural Questions (NQ) corpus \cite{47761} that consists of questions and paired Wikipedia pages, with the task of finding the exact location of the answer that is present in the article, if any. The NQ dataset is large (300k training samples) and answers (if present) can appear in either or both text and tables.
NQ contains one human annotation for every question in the training set, and 5 annotations for every question in the development and test sets. Based on the statistics of the development set, around 14\% of questions contain a short answer in a table. Only 48\% of the questions have a long answer annotation (a paragraph or a table that contains an answer), while only 33\% contain a short answer annotation (an exact location of a short answer phrase).  In this work we first evaluate our models on a subset of questions that contain at least one short answer in tables, referred as NQTables, and then further evaluate the model on the full dataset, referred as FullNQ. 

For the questions in NQTables, we further evaluate in two settings: (\textit{i}) NQTables$_{\mbox{{\small Tab}}}$, where systems are limited to predict answer spans only from tables, and (\textit{ii}) NQTables, where systems can predict answer spans from both text and tables. 
Note that although all questions in NQTables have at least one answer in a table, they might also have answers in text, and systems operating in the full NQTables setting have more chances to arrive at a correct answer than systems in NQTables$_{\mbox{{\small Tab}}}$ setting. 
While our primary focus in this work is on improving the short answer prediction, we also report the long answer prediction results for our best model used for the full NQ dataset, according to official metrics.

The development set (\textsc{Dev}) for NQTables used in this work contains 1118  questions where the short answer can be found in a table. Since the official test set of the corpus is not public, all our experiments use the official development set as our test set, while splitting the training set into training (90\%) and 2 validation sets (each contains 4\% of the data). The first validation set (\textsc{Val-1}) is used for tuning the parameters of the table-based models, while the second validation set (\textsc{Val-2}) is used for tuning the parameters to combine text-based and table-based models. For clarity, in all the experiment variations we use the notation defined above where NQTables$_{\mbox{{\small Tab}}}$-{Dev/Val-1/Val-2} is limited to predict answer spans only from tables, and NQTables-{Dev/Val-1/Val-2} is able to predict answer spans from both text and tables.

The official evaluation script computes F1 scores and considers any questions that have at least 2 annotated answers as being answerable, while questions with 1 or no answer are unanswerable. An F1 score is calculated on the ability to predict a span that matches at least one of the annotations for the answerable questions, and to correctly predict unanswerable ones. 
Since the validation sets contain only a single annotation, the F1 score measure based on 5-way annotation cannot be used directly on \textsc{Val-1} or \textsc{Val-2}. Therefore, in the table-based experiments on the \textsc{Val-1}, we report both accuracy (percent of correctly predicted answers) and modified F1 score (F1*), where modified F1 is based on the match of the predicted answer to a single annotation. Finally, we report a string-based F1 score that accepts any exact string match of a predicted span to an answer when combining text and table models. 

\subsection{Table-based Results}
\label{sec: table-based results}

\begin{table*} [t]
\begin{center}
\scalebox{.9}{
\begin{tabular}{|l|c|c|c|c|c|} \hline
 & \multicolumn{1}{c|}{\bf Pretrained on} &  & \multicolumn{2}{c|}{\textbf{\textsc{Val-1}}} & \multicolumn{1}{c|}{\textbf{\textsc{Dev}}} \\

\bf Model & \bf WikiTables & \bf Finetuned on & \bf F1* & \bf Acc & \bf F1 \\\hline
RoBERTa baseline   &  no  & FullNQ & 49.1 & 48.2 & 53.3 \\ \hline
RoBERTa tables     &  no  & tables only & 52.5 & 51.2 & 56.8  \\ \hline
\url{[SEP]} encoding       &   no & tables only & 53.2 & 51.1 & 57.5 \\ \hline
GNN &  no  & tables only & 52.6 & 51.2 & 55.0 \\ \hline
GNN &  yes &  tables only & 54.4 & 52.9 & 56.7  \\ \hline
GNN  &  yes & FullNQ $\rightarrow$ tables only & 55.7 & 54.3 & 58.9  \\ \hline
Context-aware +  GNN   &  yes & FullNQ $\rightarrow$ tables only & 56.7 & 55.4 & 60.0  \\ \hline

\end{tabular}
}
\end{center}
\caption{\label{tab: table_repr}Evaluation of the table encodings and context-aware table representation on NQTables$_{\mbox{{\small Tab}}}$-{Val$_1$} (\textsc{Val-1}) and NQTables$_{\mbox{{\small Tab}}}$-{Dev} (\textsc{Dev}) sets of NQTables that contains tables input exclusively.}
\end{table*}

First, we evaluate the proposed table representation approaches on the NQTables subset. 
In this set of experiments, we use the article's tables as our input, omitting direct usage of the article's text except in the form of context-aware updates of the model used in Section \ref{sec: qa_methods_text_aware}. 
The results on NQTables$_{\mbox{{\small Tab}}}$-{Val$_1$} and NQTables$_{\mbox{{\small Tab}}}$-{Dev} are presented in Table \ref{tab: table_repr}. First, we evaluate our model using a RoBERTa baseline, following 
\cite{alberti2019bert} with a RoBERTa pretrained model instead of BERT (line 1). This baseline is trained using the FullNQ corpus and contains both text and table inputs. We improve this baseline by finetuning on the NQTables part of the training dataset (line 2). Previous work on table representations using BERT has shown improvement from using \url{[SEP]} tokens to highlight cell boundaries \cite{hwang2019comprehensive}. This is the additional baseline we report in line 3.
Since this baseline performs well, we combine the usage of \url{[SEP]} tokens with our models for the rest of the table-encoding experiments (lines 3-7). 
We evaluate our table representation in lines 4-6, where results in lines 5 and 6 are obtained by initially pretraining the additional weights introduced by the GNN using WikiTables, as described in Section \ref{sec: pretraining}. We also found that by finetuning on the full NQ dataset first and further finetuning on the NQTables subset, the results are substantially improved (lines 6 and 7). Finally, our context-aware model combined with the table encoding from line 6 achieves the best result in this set of experiments. The improvements of the GNN and textual context-aware models are statistically significant with $p<0.01$ according to a Wilcoxon signed-rank test. In the Appendix we provide two examples --- one where context-aware model prediction improves, and the other one where adding textual context hurts.

\subsection{Combining Tables and Text}
\label{sec: experiments_text_table}

\begin{table*} [t]
\begin{center}
\scalebox{.9}{
\begin{tabular}{|l|c|c|c|c|c|c|c|c|c|c|} \hline
 & \multicolumn{3}{c|}{ \textbf {NQTables (span)}} & \multicolumn{3}{c|}{ \textbf {FullNQ (span)}} & \multicolumn{3}{c|}{ \textbf {FullNQ (str)}}\\\hline
\bf Model & \bf F1 & \bf Prec & \bf Rec & \bf F1 & \bf Prec & \bf Rec & \bf F1 & \bf Prec & \bf Rec \\\hline
\citet{alberti2019bert}\footnote{Analysis done using the model available on \url{github}.} & 58.9 & 63.6 & 54.8 & 52.4 & 57.6 & 48.0 & 54.0 & 59.3 & 49.5 \\ \hline \hline
RoBERTa   & 62.5 & 65.0 & 60.2 & 54.4 & 60.3 & 49.6 & 56.2 & 62.3 & 51.2 \\ \hline
GNN  & 63.1 & 64.6 & 61.7 & \textbf{54.6} & 60.2 & 50.0 & \textbf{56.6} & 62.3 & 51.8 \\ \hline
Context-aware + GNN  & \textbf{64.0} & 65.5 & 62.6 & \textbf{54.6} & 59.4 & 50.6 & \textbf{56.6} & 61.5 & 52.4\\ \hline

\end{tabular}
}
\end{center}
\caption{Results of combining text-based and table-based predictions from two models: F1, Precision, and Recall scores reported on NQTables-Dev and FullNQ-Dev. (span) and (str) indicate span-based and string-based scores. \label{tab: text_and_table}}
\end{table*}

In this section, we describe experiments of combining text and table predictions. Since the official development set contains up to 5 annotations, some of those annotations can be associated with tables, while others are associated with text. Unlike in the previous section, where text-related answers are ignored, 
here we allow a match to either text- or table-based answers. For this purpose, we need to combine scores from the text-based and table-based models. For text predictions, we use the RoBERTa baseline that was trained on the FullNQ dataset; for the table predictions we use our best GNN model and the context-aware model (lines 6 and 7 in Table \ref{tab: table_repr}, respectively). Oracle analysis on the development set suggests that a linear score transformation in the case of the FullNQ is not effective. To combine the scores of the text model $g_{c}$ and the table model $g_{t}$, we use grid search, 
 as described in Section \ref{sec: methods_text_table}.
The search for parameters $\alpha$, $\beta$, and $\gamma$ is done on \textsc{Val-2} containing questions with table answers, questions with text answers, and questions without answers, using the string-based F1 score instead of the span-based one to compensate for the lack of multiple annotations in the validation set. According to this metric, both answers in tables and text are considered correct if they have 
exact string match with the gold answer span in a table annotated in \textsc{Val-2}. 

We evaluate those models on both the NQTables-{Dev} and FullNQ-{Dev}, allowing both text and table answers. 
The results are presented in Table \ref{tab: text_and_table}. 
Results from our RoBERTa baseline that was trained on FullNQ are shown in line 1. Then, as mentioned above, we combine this generic model with table-only models (lines 2 and 3). Our experiments suggest that our proposed best model that uses table encoding improves the F1 score on  NQTables by 1.5 points. When the table+text model combination is used on the full NQ dataset, there is a small improvement from both table models, but the textual context attention model is comparable to the GNN model, increasing recall at the cost of reducing precision.
This might be explained by the relatively small fraction of questions that have answers only in tables (8\% of all questions) and the difficulty in calibrating answers from tables and text against each other. 

\subsubsection{Impact of Relation Types}
\begin{table} [t]
\begin{center}
\begin{tabular}{|l|c|c|} \hline
 & \multicolumn{2}{c|}{ \textbf{\textsc{Val-1}}} \\
  & \textbf{F1*} & \textbf{Acc}\\ \hline
all relations & 55.7 & 54.3 \\ \hline
$-$ cell-col  & 54.9 & 53.7\\ \hline
$-$ cell-row  & 55.0 & 53.7\\ \hline
$-$ in-cell   & 56.0 & 53.8\\ \hline
$-$ cross-col & 55.7 & 54.3\\ \hline
$-$ cross-row & 55.7 & 54.7\\ \hline
\end{tabular}
\end{center}
\caption{Importance of relation types.\label{tab: rel_types}}
\end{table}

In order to investigate which of the five types of table relations (defined in Section \ref{sec: relations}) help table representation the most we perform an ablation study where we use the baseline of the pretrained GNN model (line 6), but remove each type of relation. The results are presented in Table \ref{tab: rel_types}. The ablation study shows a clear importance of cell-column and cell-row relation types, where the performance of the model without each of those relations degrades. While the F1 score for the experiment without in-cell relation is higher, the accuracy is much lower. The cross-column relation does not seem to contribute to the overall performance while the cross-row relation degrades the accuracy. 

\subsubsection{Impact of Negative Sampling}

\begin{table*} [t]
\begin{center}
\scalebox{.9}{
\begin{tabular}{|l|c|c|c|} \hline
  & \textbf {NQTables } & \textbf {FullNQ } & \\
\bf Sampling ( pos : neg\_within : neg\_outside) & \textbf {\textsc{Val-1} (F1)} & \textbf {\textsc{Dev} (F1)} & \bf $\gamma$ \\\hline
(1) Equal sampling within positive article (1:1:0)  & 55.7 & 54.6 & -0.005\\ \hline
(2) Random sampling across all articles (0.63:0.28:0.72) & 52.5 & 55.1 & -3 \\ \hline
(3) Equal sampling within and across articles (2:1:1) & 54.0 & 54.9 & -0.5 \\ \hline
\end{tabular}
}
\end{center}
\caption{The effect of negative sampling technique on table-only models and text+table model combinations.\label{tab: ablation}}
\end{table*}

In all our experiments using models finetuned on  tables only data (lines 2-7 in Table \ref{tab: table_repr} and lines 2-3 in Table \ref{tab: text_and_table}), during training we used an equal proportion of positive and negative samples,\footnote{A sample is a table or a table fragment, together with an indication of short answer span or \textsc{null}.} where all  negative samples were taken from the articles in NQTables-Train, which contain an answer in a table. This approach is successful when models were evaluated on their ability to predict answers in tables, for articles that are known to contain such answers.
\commentout{This was done based on our initial experimentation that showed the benefit of such approach for table-based experiments (Section \ref{sec: table-based results}) comparing to random sampling of negatives across all the articles, where the ratio of negatives-to-positives is 0.63, as used in \citet{alberti2019bert}.} On the other hand, when the table-based models are asked to make predictions for articles not known to contain answers in tables (or any answers), they tended to be over-confident in comparison to a generic text-based model, trained with negative examples across all articles. This over-confidence is evident from the extremely high selected confidence parameters of
$\gamma=-0.005$ for GNN model (line 2 in Table \ref{tab: text_and_table}) and $\gamma=-0.0025$ for context-aware model (line 3 in Table \ref{tab: text_and_table}),
suggesting that only high-value and high-confidence scores are considered from the table-based model.  

To investigate the effect of the negative sampling method we perform an ablation study that compares three techniques: 1) sampling negative samples from within  articles that contain an answer in tables, 2) random sampling of negatives across all NQ articles, in the proportion used by  \citet{alberti2019bert}, and 3) sampling negatives with equal proportion from articles that contain answer in tables, and articles that do not. The results for the GNN model are shown in Table \ref{tab: ablation}. While the first sampling strategy works best when a table-based model is used to predict answers from tables in NQTables, sampling negatives from a more diverse set of articles improves the overall FullNQ results for the combination of text and table-based models, by allowing better calibration between text and table models. The threshold values $\gamma$ are seen to be much lower in these cases. However, the second and third strategies reduce performance on QA from tables. When optimizing the random sampling strategy for highest performance on FullNQ, no benefit was found from contextual text attention for table representations.


Finally, we compare the performance of our model to other work on the Natural Questions on both the short and long answer prediction tasks.
In our model, the long answer is predicted based on the segment that corresponds to the short answer prediction. The results are presented in Table \ref{tab: related}. As we can see, our method obtains a substantial improvement over the baseline  of \newcite{alberti2019bert} in short answer F1, and a smaller improvement in long answer F1.  Advances in long answer F1 from state-of-the-art recent works are likely complementary to our method and can be integrated for additive gain.
\begin{table} [t]
\begin{center}
\scalebox{.9}{
\begin{tabular}{|l|c|c|} \hline
\bf Model & Short F1 & Long F1 \\\hline
\citet{alberti2019bert} & 52.7 & 64.7 \\ \hline 
\citet{liu2020rikinet} & 57.7 & 73.9 \\ \hline
\citet{ravula2020etc} & 58.5 & 78.2 \\ \hline
GNN, random sample (2) & 55.1 & 65.9 \\ \hline
\end{tabular}
}
\end{center}
\caption{Comparison to other NQ models.\label{tab: related}}
\end{table}

\section{Conclusion}

Tables in Web documents are pervasive and can be directly used to answer many search queries.  In this work, we presented an approach to enrich table representations using information from article text, and showed that it improves a state-of-the-art pretrained structure-aware table representation for question answering from tables. We also studied how to effectively combine text-based and table-based approaches.
Finally, we performed the first study focusing on table QA for the Natural Questions dataset, and showed that improved representations of tables lead to performance gains.
%
In future work, our methods can be applied to other QA datasets, such as WikiTableQuestions \cite{PasupatLiang15} and HybridQA \cite{chen2020hybridqa}.


\section*{Acknowledgments}
This work was supported by the Google Faculty Research Awards Program.  Computational resources were provided by the GCP research credits program. We thank the reviewers for their helpful feedback.

\bibliographystyle{acl_natbib}
\bibliography{emnlp2020}

\clearpage
\appendix
\onecolumn
\section{Examples}

\begin{figure}[h]
    \begin{center}
    \includegraphics[width=0.9\textwidth]{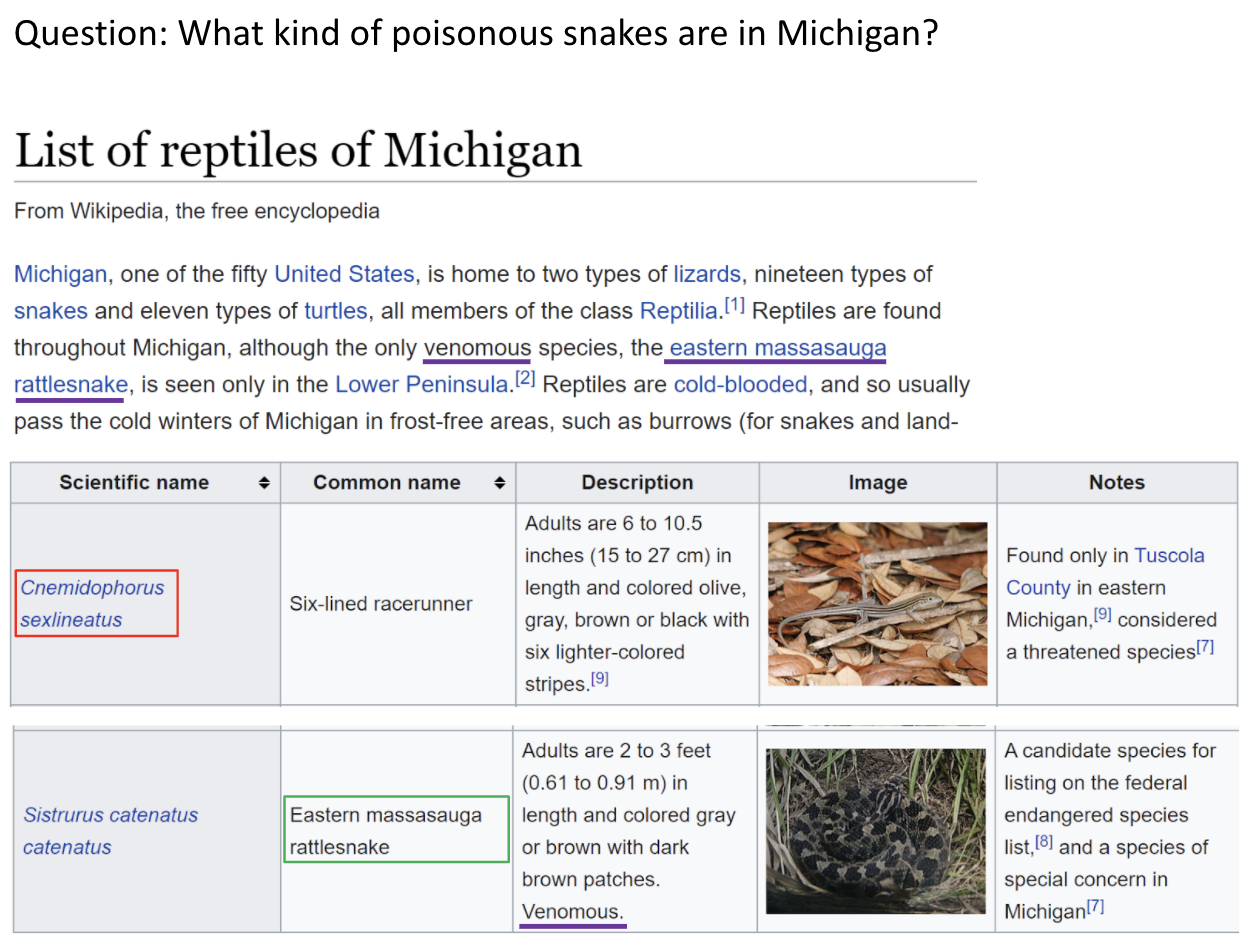}    
    \end{center}
    \caption{Example where incorporating textual context helps. In red square the prediction was made by GNN model without context attention (Table \ref{tab: table_repr}, line 6), in green square the prediction was made by context-aware GNN (Table \ref{tab: table_repr}, line 7).}
    \label{fig: example_snake}
\end{figure}
The correct answer "Easter massasauga rattlesnake" was correctly predicted from the table when context-aware attention was used. The same answer can be extracted based on the textual information in the introduction paragraph. By propagating relevant information from the textual context to the table entries, the model was able to predict correctly the answer from the table. On the other hand, the table-only model which does not use the surrounding textual context incorrectly predicted "Cnemidophorus sexlineatus".

\clearpage
\begin{figure}[h]
    \begin{center}
    \includegraphics[width=0.9\textwidth]{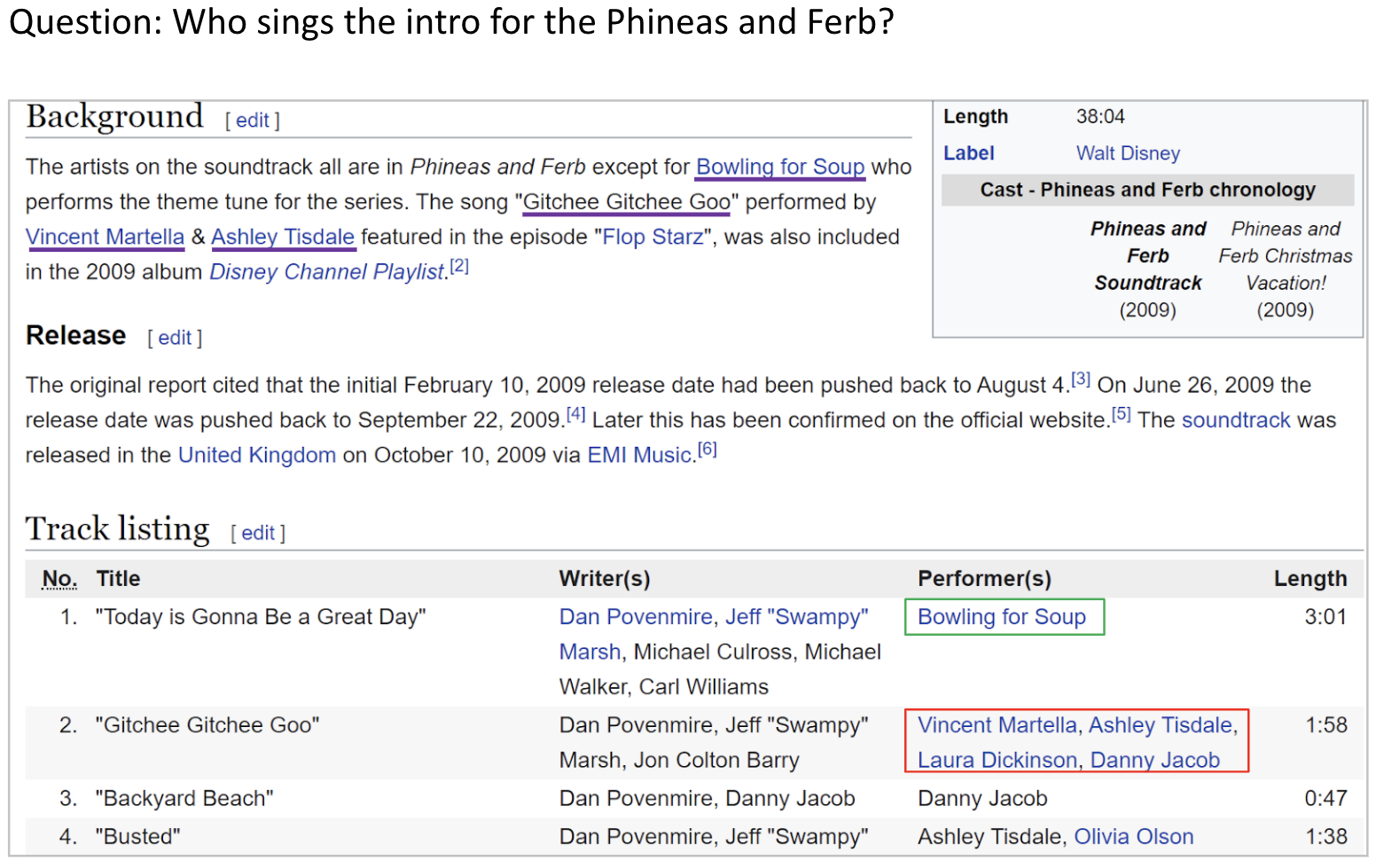}    
    \end{center}
    \caption{Example where incorporating textual context hurts. In red square the incorrect prediction was made by context-aware GNN model (Table \ref{tab: table_repr}, line 7), in green square the correct prediction was made by GNN without context attention (Table \ref{tab: table_repr}, line 6).}
    \label{fig: example_song}
\end{figure}
The correct answer "Bowling for Soup" was correctly predicted from the table by the model that does not use context attention, while the context-aware model predicts the answer that has higher number of links to the textual context.

\end{document}